%% file: neurips_2022.tex
\documentclass{article}

\usepackage[preprint]{neurips_2022}

\usepackage[utf8]{inputenc} 
\usepackage[T1]{fontenc}    
\usepackage{hyperref}       
\usepackage{url}            
\usepackage{booktabs}       
\usepackage{amsfonts}       
\usepackage{nicefrac}       
\usepackage{microtype}      
\usepackage{xcolor}         
\usepackage{adjustbox}

\usepackage{subcaption}
\usepackage{graphicx}
\usepackage{amsmath}

\title{Compressing Pre-trained Transformers via Low-Bit NxM Sparsity for Natural Language Understanding}

\input{macro}

\author{%
  Connor Holmes\\
  Colorado School of Mines\\
  Golden, CO 80401 \\
  \texttt{cholmes@mines.edu}
  \And
  Minjia Zhang\\
  Microsoft\\
  Bellevue, WA 98004\\
  \texttt{minjiaz@microsoft.com}
  \AND
  Yuxiong He\\
  Microsoft\\
  Bellevue WA, 98004\\
  \texttt{yuxhe@microsoft.com}
  \And
  Bo Wu\\
  Colorado School of Mines\\
  Golden, CO 80401\\
  \texttt{bwu@mines.edu}
}

\begin{document}

\maketitle

\begin{abstract}
In recent years, large pre-trained Transformer networks have demonstrated dramatic improvements in many natural language understanding tasks. However, the huge size of these models brings significant challenges to their fine-tuning and online deployment due to latency and cost constraints. New hardware supporting both N:M semi-structured sparsity and low-precision integer computation is a promising solution to boost DNN model serving efficiency. However, there have been very few studies that systematically investigate to what extent pre-trained Transformer networks benefit from the combination of these techniques, as well as how to best compress each component of the Transformer. We propose a flexible compression framework \ours that performs simultaneous sparsification and quantization using ADMM and STE-based QAT. Furthermore, we present an inexpensive, heuristic-driven search algorithm that identifies promising heterogeneous compression configurations that meet a compression ratio constraint. When evaluated across the GLUE suite of NLU benchmarks, our approach can achieve up to 93\% compression of the encoders of a BERT model while retaining 98.2\% of the original model accuracy and taking full advantage of the hardware’s capabilities. Heterogeneous configurations found by the search heuristic maintain 99.5\% of baseline accuracy while still compressing the model by 87.5\%.
\end{abstract}

\input{intro}

\input{background}

\input{design}

\input{eval}

\input{limitations}

\input{conclusion}

\begin{ack}
Use unnumbered first level headings for the acknowledgments. All acknowledgments
go at the end of the paper before the list of references. Moreover, you are required to declare
funding (financial activities supporting the submitted work) and competing interests (related financial activities outside the submitted work).
More information about this disclosure can be found at: \url{https://neurips.cc/Conferences/2022/PaperInformation/FundingDisclosure}.

Do {\bf not} include this section in the anonymized submission, only in the final paper. You can use the \texttt{ack} environment provided in the style file to autmoatically hide this section in the anonymized submission.
\end{ack}

\bibliography{reference}

\section*{Checklist}

The checklist follows the references.  Please
read the checklist guidelines carefully for information on how to answer these
questions.  For each question, change the default \answerTODO{} to \answerYes{},
\answerNo{}, or \answerNA{}.  You are strongly encouraged to include a {\bf
justification to your answer}, either by referencing the appropriate section of
your paper or providing a brief inline description.  For example:
\begin{itemize}
  \item Did you include the license to the code and datasets? \answerYes{See .}
  \item Did you include the license to the code and datasets? \answerNo{The code and the data are proprietary.}
  \item Did you include the license to the code and datasets? \answerNA{}
\end{itemize}
Please do not modify the questions and only use the provided macros for your
answers.  Note that the Checklist section does not count towards the page
limit.  In your paper, please delete this instructions block and only keep the
Checklist section heading above along with the questions/answers below.

\begin{enumerate}

\item For all authors...
\begin{enumerate}
  \item Do the main claims made in the abstract and introduction accurately reflect the paper's contributions and scope?
    \answerYes
  \item Did you describe the limitations of your work?
    \answerYes{See Section \ref{sec:limitations}}
  \item Did you discuss any potential negative societal impacts of your work?
    \answerYes{See Section \ref{sec:limitations}}
  \item Have you read the ethics review guidelines and ensured that your paper conforms to them?
    \answerYes
\end{enumerate}

\item If you are including theoretical results...
\begin{enumerate}
  \item Did you state the full set of assumptions of all theoretical results?
    \answerNA{}
        \item Did you include complete proofs of all theoretical results?
    \answerNA{}
\end{enumerate}

\item If you ran experiments...
\begin{enumerate}
  \item Did you include the code, data, and instructions needed to reproduce the main experimental results (either in the supplemental material or as a URL)?
    \answerYes{Code and instructions for reproducing main experimental results is included in the supplementary materials.}
  \item Did you specify all the training details (e.g., data splits, hyperparameters, how they were chosen)?
    \answerYes{See Section \ref{sec:evaluation}}
        \item Did you report error bars (e.g., with respect to the random seed after running experiments multiple times)?
    \answerNo
        \item Did you include the total amount of compute and the type of resources used (e.g., type of GPUs, internal cluster, or cloud provider)?
    \answerYes
\end{enumerate}

\item If you are using existing assets (e.g., code, data, models) or curating/releasing new assets...
\begin{enumerate}
  \item If your work uses existing assets, did you cite the creators?
    \answerYes{See Section ~\ref{sec:evaluation}}
  \item Did you mention the license of the assets?
    \answerYes{See Section ~\ref{sec:evaluation}}
  \item Did you include any new assets either in the supplemental material or as a URL?
    \answerYes{Code for \ours is included as supplementary material}
  \item Did you discuss whether and how consent was obtained from people whose data you're using/curating?
    \answerNA
  \item Did you discuss whether the data you are using/curating contains personally identifiable information or offensive content?
    \answerNA
\end{enumerate}

\item If you used crowdsourcing or conducted research with human subjects...
\begin{enumerate}
  \item Did you include the full text of instructions given to participants and screenshots, if applicable?
    \answerNA
  \item Did you describe any potential participant risks, with links to Institutional Review Board (IRB) approvals, if applicable?
    \answerNA
  \item Did you include the estimated hourly wage paid to participants and the total amount spent on participant compensation?
    \answerNA
\end{enumerate}

\end{enumerate}

\newpage
\appendix

\section{Appendix}

\input{appendix}

\end{document}

%% file: macro.tex
\usepackage{xspace}

\newcommand{\originalgrumbler}[2]{\begin{quote}\textcolor{blue}{\sl{\bf #1 says:} #2}\end{quote}}
\newcommand{\grumbler}[2]{\originalgrumbler{#1}{#2}}
\newcommand{\minjia}[1]{\grumbler{Minjia}{#1}}

\newcommand{\connor}[1]{\grumbler{Connor}{#1}}

\newcommand{\name}{NxMiFormer\xspace}
\newcommand{\ours}{NxMiFormer\xspace}



%% file: intro.tex
\section{Introduction}
\label{sec:intro}

Large-scale Transformer-based models, such as BERT~\citep{bert}, RoBERTa~\cite{roberta}, and T5~\citep{t5}, have achieved outstanding performance for a wide variety of natural language tasks, such as natural language inference~\citep{xlnet,t5}, question answering~\citep{roberta}, and others. 
However, these models raise significant challenges in deployment due to latency and cost constraints. Various techniques have been proposed to reduce  the number of parameters as well arithmetic operations for these models, including but not limited to knowledge distillation~\citep{distilbert,minilm,tinybert,mobilebert,ladabert}, sparsification~\citep{lottery-bert,proximal-pruning-bert, nxmtransformer}, quantization~\citep{q8bert,q-bert,binary-bert}, and neural architecture search~\citep{adabert}. However, how to effectively compress Transformer models remain an open and challenging research problem, because running models at low latency and cost has always been extremely desirable. 


Sparse Tensor Core (STC) has entered the realm of DNN acceleration since NVIDIA released the Ampere architecture~\citep{nvidia-nxm}. 
STC supports acceleration of compressed model in two ways: (1) the support of N:M semi-structured sparsity, which allows accelerated execution when the model weights contain at most M non-zero parameters out of N consecutive parameters, and (2) low-bit integer arithmetic computation, which allows accelerating operations with INT8/INT4 values. 



There are already existing efforts that allow one to reap the performance benefit of STC. To benefit from low-bit computation, one could quantize model weights and activations into INT8/INT4 using popular quantization methods, such as post-training quantization~\cite{ptq} and quantization-aware training (QAT)~\cite{qat}. The former directly replace the floating point values of a trained model with a low-precision representation, like using a symmetric scalar quantizer. The latter quantizes the network during training to improve the model accuracy from quantization errors. Multiple extension of QAT have also been introduced to further push the quantization precision from FP32 to pure INT8 or mixed INT8/INT4 precision for BERT~\citep{q8bert,q-bert,hawq-v2,i-bert}. On the sparsification side, pruning methods have shown to be extremely effective at reducing the number of parameters. Approaches such as magnitude pruning~\cite{deep-compression-eie}, L0 regularization~\cite{l2-regularization}, the lottery ticket hypothesis~\cite{lottery-hypothesis}, and movement pruning~\cite{movement-pruning} have demonstrated remarkable reductions in model size. However, these methods yield little actual efficiency benefits on STC because their sparsity is still too irregular to satisfy the N:M sparsity hardware constraints. More recently, \cite{nxmtransformer} investigates how to induce N:M sparsity via ADMM, leading to promising accuracy results. However, none of the above have explicitly considered combining N:M sparsity and low-bit computation to maximize the efficiency gains on STC.

Beyond the quantization and sparsification schemes, compressing Transformer language models incurs yet another dimension of difficulty  ---  state-of-the-art transformer models are trained in two stages: a pre-training stage on open-domain web text data followed by an adaptation stage that transfers the knowledge to task-specific domains. Given this transfer learning paradigm, the compression for Transformer models also fall into two major categories: pre-training stage compression and adaptation stage compression. The pre-training stage often presumes compression constraints at model instantiation. For example, in the case of training a Transformer model from scratch with a fraction of its weights already frozen to 0~\citep{sr-ste}. In contrast, the adaptation stage compression would compress with a fully pre-trained dense model~\citep{movement-pruning,nxmtransformer}. While pre-training compression can achieve higher accuracy when combined together with adaptation compression~\citep{light-paff}, the adaptation stage compression in isolation is much cheaper than pre-training compression as the adaptation stage often learns on a much smaller dataset.

In this work, we demonstrate N:M semi-structured sparsity and quantization may be jointly induced using adaptation stage compression in pre-trained Transformer networks. Furthermore, we investigate how different compression techniques may best be combined in order to maximize accuracy under a given compression target.
We make the following contributions:
(1) We present a flexible compression framework, \name, for inducing N:M semi-structured sparsity and low-bit quantization in Transformer-based models that using a hybrid ADMM-STE approach.
(2) We investigate what approach to quantization best preserves accuracy along-side ADMM-based sparsification.
(3) We demonstrate the combination of N:M semi-structured sparsity can realize end-to-end inference improvements.
(4) We develop an inexpensive heuristic-based search algorithm to navigate the compression space of combined N:M semi-structured sparsity and low-bit quantization.

%% file: background.tex
\section{Background and Related Work}
\label{sec:related}

\textbf{Compression Techniques for Pre-trained Transformers.} 
Many works are dedicated to compressing large-scale pre-trained language models. Among them, sparsification has been demonstrated to be a successful technique for achieving a high compression ratio. \citet{lottery-bert} extend the Lottery Ticket Hypothesis \citep{lottery-hypothesis} to pre-trained models, finding that winning tickets for the pre-training tasks may transfer universally to downstream tasks. \citet{reweighted-proximal-pruning} show that a proximal pruning strategy achieves higher accuracy than competing lasso regularization methods and iterative magnitude pruning. However, most of these studies focus on unstructured sparsity, which encounters difficulty in obtaining large speedups on modern hardware due to the irregularity in the patterns of retained weights. On the other hand, \citet{sixteen-heads,block-pruning-transformer} observe that many transformer heads themselves are redundant and can be pruned with minimal accuracy loss. Despite having a structured weight pattern, these techniques cannot take advantage of hardware that supports semi-structured sparsity. Furthermore, these techniques evaluate sparsification in isolation and do not investigate the impact of low-width integer quantization.

Pre-trained Transformers, such as BERT, can also realize compression gains and increased inference efficiency with quantization ~\citep{q8bert,q-bert,hawq-v2,i-bert}. Prior work even shows the feasibility of only binary weights in the extreme case~\citep{binary-bert}. Since Transformer networks often have high memory footprints and require significant compute and bandwidth resources during inference, quantization not only reduces the model size but also accelerate inference time with reduced memory bandwidth consumption and compatibility with accelerators, such as the dense Tensor Cores.  However, inference benefits diminish as bit-width decreases since models incur unacceptable accuracy when the intermediate activations are quantized below 8-bits; the low-width weights are then also treated by the ALUs as 8-bit integers and maintain the same compute efficiency. Unlike these quantization frameworks, we consider quantization along-side N:M semi-structured sparsity.

\noindent
\textbf{Sparse Tensor Core.} Hardware (e.g., NVIDIA A100 GPU) supporting Sparse Tensor Core operations has become increasingly availble. 
There have been a limited number of studies that aim to compress DNN models to better leverage this hardware support for increased inference performance. To obtain N:M sparsity, ASP \citep{nvidia-nxm} proposes training the dense network until convergence and then using single-shot magnitude-based pruning (i.e., with a fixed mask) to obtain sparsity conformant to the N:M constraints. It then requires repeating the original training procedure to recover accuracy. \citet{sr-ste} propose to train a model from scratch with an N:M mask, using a sparse-refined straight-through estimator (SR-STE). However, SR-STE sparsifies from the random model initialization, which avoids the costly sparse retraining but also necessitates performing the pre-training process with only a sparse representation in mind. As compression-aware approaches, these approaches provide high accuracy but cannot be efficiently used to explore the compression space cheaply and may exceed the resources available for all but the largest organizations. \citet{nxmtransformer} investigates how to best induce floating-point semi-structured sparsity during the adaptation phase using Alternating Directions Method of Multipliers (ADMM). However, there is almost no research that has been published in combining the two and assessing their impact on NLU tasks.

\noindent
\textbf{ADMM for Neural Networks.} 
The alternating direction method of multipliers (ADMM)~\citep{admm} has been extensively studied for constrained optimization problems, by breaking them into smaller pieces. Prior work uses ADMM as the sparsifying and/or quantization mechanism for model compression for computer vision tasks~\citet{progressive-admm,admm-nn,unified-admm,pconv,patdnn,differentiable-admm,automatic-admm}.There has been limited studies on how ADMM can benefit the new STC hardware and pre-trained Transformer-based language models. As the only existing work to focus on natural language models, NxMTransformer ~\citep{nxmtransformer} is the most closely related work. NxMTransformer uses ADMM to induce N:M semi-structured sparsity on Transformer models; however, NxMTransformer does not consider the impact of quantization at all, only evaluating model accuracy for floating point sparsity.

%% file: design.tex
\section{\ours: Compressing Transformers via Low-bit N:M Sparsity} \label{sec:comp_methodology}

In this section, we formulate our combined optimization problem, identify appropriate sparsification and quantization techniques, and note how to combine them into a single compression framework.

\subsection{Problem Formulation}

In this work, we consider a Transformer-based language model $\Phi(W)$ with a collection of $L$ weights $W=\{\mathbf{W}_i\}_{i=1}^{L}$.
The model is first pre-trained on open-domain data to get $W^0=\{\mathbf{W^0}_i\}_{i=1}^{L}$ and then adapts to $\{W'_i\}_{i=1}^L$ with a domain-specific dataset $D$ for a natural language understanding task, such as sentiment analysis, entailment, question-answering, etc. $D$ is composed of input (e.g., text phrases) and target pairs: $\{x_i,y_i\}_{i=1}^N$. The goal of \ours is to load $W^0$ and fine-tune it on $D$ to $W'$ such that each ${W'}_i \in W'$ can satisfy both N:M semi-structured sparsity and quantization constraints simultaneously. Finally, $\Phi(W')$ should achieve similar performance in comparison to fine-tuning the task-specific objective function $f(\{\mathbf{W}_i\}_{i=1}^{L})$ using $W_0$ but without the constraints.

\subsection{Sparsification Approach}
\label{subsec:sparsification_approach}

Given its effectiveness in NxMTransformer \citep{nxmtransformer}, \ours adopts Alternating Directions Method of Multipliers as its mechanism for inducing semi-structured sparsity. ADMM decomposes the problem of sparsification, which itself is non-convex with combinatorial constraints, into two subproblems: one accuracy promoting ( Eq. \ref{eq_first-subproblem}) and one sparsity promoting (Eq. \ref{eq_second-subproblem}).

\begin{equation} \label{eq_first-subproblem}
	\min_{\{\mathbf{W}_i\}} f(\{\mathbf{W}_i\}_{i=1}^{L}) + 
	\sum_{i=1}^{L} \frac{\rho}{2} \Vert \mathbf{W}_i - \mathbf{Z}_{i}^k + \mathbf{U}_{i}^k \Vert_{F}^{2} 
\end{equation}

\begin{equation} \label{eq_second-subproblem}
	\min_{\{\mathbf{Z}_i\}} \sum_{i=1}^{L} g_i(\mathbf{Z}_i) + 
	\sum_{i=1}^{L} \frac{\rho}{2} \Vert \mathbf{W}_{i}^{k+1} - \mathbf{Z}_{i} + \mathbf{U}_{i}^k \Vert_{F}^{2} 
\end{equation}

\begin{equation} \label{eq:admm-projection}
	\mathbf{Z}_{i}^{k+1} = \Pi_{S_i}(\mathbf{W}_{i}^{k+1} + \mathbf{U}_{i}^{k})
\end{equation}

The accuracy promoting subproblem can be solved (e.g., via Adam~\citep{adam}) with the same complexity of training $f(\cdot)$. In the second subproblem, the indicator function $g_i(\cdot)$ is 0 if our constraint is met; this subproblem then collapses to the Euclidean projection of $\mathbf{W}_{i}^{k+1} + \mathbf{U}_{i}^{k}$ onto sparsity constraint (Equation ~\ref{eq:admm-projection}). This can be solved trivially by retaining the $M$ largest values out of each contiguous group of $N$. See the Appendix for further information on decomposition via ADMM.

\subsection{Quantization Approaches}
\label{subsec:quantization_approaches}

For integer quantization, we formulate a matrix as one whose values are represented as:

\begin{equation}
    x_d = s_i * (x_q + o_i) 
    \label{eqn-quantization}
\end{equation}

where $x_d$ is the encoded value, $s_i$ is the per-group scaling factor, $o_i$ is the per-layer offset, and $x_q$ is the encoded integer representation. Since symmetric quantization, where $o_i = 0$, lends itself to easier runtime implementations, we focus on this configuration. Under this scheme, a floating-point value may be quantized as in Eqn.~\ref{eqn:symmetric-quant}, where it is divided by the encoding scale and then clamped to the range of the integer representation we are converting.

\begin{equation} \label{eqn:symmetric-quant}
x_q = clamp(round(\frac{x}{s}), -2^{b-1} + o, 2^{b-1} - 1 + o)
\end{equation}

To investigate how to best induce low-bit NxM sparsity, we consider both ADMM-based quantization methods and a Straight-through Estimator (STE) based QAT approach. The ADMM-based approaches integrate seamlessly into the same training pipeline used for quantization, while STE-based QAT provides state-of-the-art accuracy for standalone quantization ~\citep{q-bert}. Both approaches support group-wise quantization, in which subsets of neurons within a single weight matrix are quantized using different scales. For runtime simplicity, parameters within the same neuron are always quantized with the same scale.

\subsubsection{Quantization by ADMM}
\label{subsubsec:admm_quant}

Inducing a quantized model via ADMM happens in much the same was as with semi-structured sparsity. The indicator function $g_i(\cdot)$ now refers to whether a weight matrix can be quantized as in Equation \ref{eqn-quantization}. Unlike in N:M semi-structured sparsity, there is an additional degree of flexibility in choosing $s_i$. We develop two approaches for choosing the scale to compress with ADMM: minimizing the total quantization distance and minimizing the quantization error of the largest magnitude parameter.

\textbf{Minimize Error on the Largest Parameter.}
Minimizing error on the largest parameter is the ADMM analog to the traditional quantization approach. The scale $s_i$ for each group of weights is set to the largest magnitude element of the sum $\mathbf{W}_{i}^{k+1} + \mathbf{U}_{i}^{k}$ divided by the largest integer representation. This approach will be referred to as ADMM-Max.

\textbf{Minimize Quantization Distance.}
This approach chooses the scale such that it minimizes the Euclidean norm between the quantized weight matrix and its raw representation. This is an "equitable" approach in that it considers the impact of quantization on more parameters than just the largest parameter. Thisdoes not guarantee better accuracy though since smaller (and by extension less important) parameters are considered to have the same sensitivity to quantization as the largest parameters. This approach will be referred to as ADMM-Dist.

To combine with quantization and N:M semi-structured sparsity within ADMM, we treat the single indicator function $g_i(\cdot)$ as the product of two indicator functions, $g^s_i(\cdot)$ and $g^q_i(\cdot)$, which correspond to our sparsity and quantization constraints respectively. The Euclidean projection must now be solved such that both constraints are met. In practice, this can be done optimally by solving the individual constraints in sequence. While the order has no effect when using ADMM-Max (since the largest parameter is preserved by the NxM semi-structured sparsity), ADMM-Dist will achieve better results when the sparsity projection is solved first.

\subsubsection{Quantization by STE-based QAT}
\label{subsubsec:ste_quant}

STE-based QAT has demonstrated state-of-the-art accuracy for standalone quantization. Since the quantization operation itself is non-differentiable, we use an STE \citep{ste} to backpropagate gradients through our quantization operator. 

ADMM-based sparsification and STE-QAT largely act orthogonally from the optimizer's perspective. ADMM sparsifies via backpropagation on the regularizer between a weight and its auxiliary variable, while STE-QAT will quantize from the original training loss function. In practice, we find it to be a reasonable approach to have both ADMM and STE-QAT running alongside each other with no additional considerations.

\section{Searching the Heterogeneous Compression Space}
\label{sec:methodology_search}

A heterogeneous compression scheme, in which the different components of each encoder may be compressed differently, can improve both accuracy and performance for a given compression ratio by better allocating compute and storage resources to different components of the encoder according to their error sensitivities. The combination of semi-structured sparsity and low-bit integer quantization provides additional flexibility towards finding an ideal compression scheme. However, exploring the combined search space of both techniques, even when using the same scheme for each layer of the model, would be prohibitively costly from a resource standpoint.

We propose evaluating a compression scheme $C$ by immediately constraining a model $M$ to its compression constraints $M'_C$ and evaluating it on a validation dataset $V$:

\begin{equation}
    \label{heuristic}
    H(C) = f(M'_C|V)
\end{equation}

While the raw loss (and corresponding accuracy) of this approach is severely degraded compared to the configuration as compressed by \ours, we find that the relative accuracy of configurations is accurately captured by this heuristic. In a sense, it measures how the optimization distance \ours will need to close in order to maintain the original model's accuracy; configurations with smaller distances are able to more easily recover the accuracy lost during the compression stage.

To heterogeneously compress a model, our system evaluates all compression configurations that meet a compression ratio constraint (e.g., 88\% compression of the encoder stack) with our compression proxy heuristic. The top-K configurations as evaluated by the heuristic are tracked, and the configuration with the highest expected runtime efficiency (as measured by effective FLOP reduction) is compressed using \ours. In this usage, increasing K will tend to increase runtime acceleration while potentially decreasing accuracy; a larger K means that we tolerate a lower heuristic score and greater compression difficulty. See Section \ref{subsec:eval_search} for the performance of this heuristic.

We also note that we are able to finish the search in a relatively short amount of time (e.g., finding a configuration for CoLA takes under an hour using a single V100), because the compression proxy introduced in Equation~\ref{heuristic} only takes a short amount of time for each compression scheme and the overall search space is still quite manageable (e.g., 905 compression configurations are evaluated at a constraint of 87.5\%). \ours can be combined with more advanced search algorithms to further accelerate the search speed of the heterogeneous compression space, especially when the model becomes larger and deeper. We leave this exploration to future work.

%% file: eval.tex
\section{Evaluation and Analysis}
\label{sec:evaluation}

In this section, we evaluate the following:
(1) How does \ours compare to the state-of-the-art techniques for inducing semi-structured sparsity and quantization on their own?
(2) Which quantization approach is the best to combine with ADMM-based sparsification?
(3) What performance gains may be realized by adopting both semi-structured sparsity and integer quantization for inference?
(4) What heterogeneous encoder designs does \ours-Search heuristic discover?

\textbf{Implementation.}
We implement \ours as a Pytorch \citep{pytorch} compatible library for flexible, heterogeneous compression. \ours performs module-level replacement of model components to target specific portions of the model with user-controlled sparsification schemes. Additionally, we provide a HuggingFace Transformers \footnote{Licensed under Apache 2.0} \citep{transformers-huggingface} compatible Trainer to enable straightforward integration with their model collection and training scripts. \ours includes built-in support for ADMM-based compression methods and quantization via STE. \ours supports arbitrary N:M sparse patterns (eg, 4:1, 8:4) so long as the weight's outer dimension is an even multiple of N. Furthermore, arbitrary integer widths for compression are supported. We evaluate 4:2 semi-structured sparsity and both 4-bit and 8-bit symmetric integer quantization due to their support in commodity hardware. Unless specified otherwise, \ours quantizes each group of 32 neurons in a weight matrix with its own scale. For all integer-quantized models, \ours uses 8 bits as further compression leads to unacceptable accuracy loss.

All fine-tuned models are derived from the pre-trained model checkpoint for BERT\footnote{\url{https://huggingface.co/bert-base-uncased}} provided by the HuggingFace model repository. PyTorch 1.8 was used alongside Transformers 4.9.1. The compression cost of \ours is equivalent to performing a second iteration of adaptation, with \ours introducing negligible overhead in the training process. Experiments were run on an internal cluster with a mix of NVIDIA V100 and A100 GPUs.

\textbf{Dataset.}
We evaluate \ours and our baselines using the General Language Understanding Evaluation (GLUE) benchmark \citep{glue}, a collection of NLP tasks varying in data availability and complexity. We report the Spearman correlation for STS-B, the F1 score for MRPC, Matthews correlation for CoLA, and accuracy for all remaining tasks. The reported average is the arithmetic mean of reported scores.

\textbf{Hyperparameters.}
In \citet{bert}, the authors only report the development results on a few tasks. Therefore, we produce the BERT baseline results independently. We perform a grid search of batch sizes 16 and 32, learning rates $\{$1e-5, 3e-5, 5e-5$\}$ for all configurations.  We follow \citet{bert} to set all other training hyperparameters. Our BERT baseline results are comparable to the results reported in the original paper. For compression via \ours, we fine-tune BERT for 10 epochs on each downstream task at a batch size of 16 across learning rates $\{$5e-5, 7e-5, 1e-4$\}$ while additionally tuning the penalty coefficient $\rho = \{$1e-3, 4e-3, 1e-2$\}$. We report the median of 5 runs of different random seeds of selected configuration on the validation set. 

To evaluate the effectiveness of \ours, we compare with the following baselines:

\begin{itemize}
    \item \textbf{BERT}~\citep{bert}: This is the BERT\textsubscript{base} model from publicly available checkpoint. This model is dense and at full 32-bit floating point precision.
    \item \textbf{NxMTransformer} ~\citep{nxmtransformer}: NxMTransformer is a compression method that acts concurrently with the adaptation stage and is the previous state of the art for N:M compression without performing sparse pre-training. NxMTransformer does not support quantization and cannot achieve the same compression ratios or inference efficiency as \ours.
    \item \textbf{Q-Bert} ~\citep{q-bert}: Q-Bert demonstrates state-of-the-art accuracy for low-integer width compression. 
    \item \textbf{\ours}: This is our method for flexible, fused compression as described in Section~\ref{sec:comp_methodology}. Unless specified otherwise, \ours uses the STE-based QAT approach along-side ADMM-based sparsification. See Section~\ref{subsec:eval_quantization} for the implications of each choice for quantization.
\end{itemize}

\subsection{GLUE Results}
\label{subsec:analysis_results}

Table ~\ref{tab:glue_results} reports the results of \ours on the GLUE baseline. While achieving a compression ratio of 93.75\% for the encoder stack, \ours retains 98.2\% of the accuracy of the dense baseline. As compared to the Q-Bert baseline, \ours is within 0.6 points for both tasks reported while achieving a further 25\% compression of the model stack and enabling higher inference throughput through the support of sparsity. \ours retains 99\% percent of the accuracy of NxMTransformer for the reported tasks and actually outperforms NxMTransformer on QNLI and both MNLI tasks. The ability of \ours to outperform NxMTransformer while simultaneously adding integer quantization support is due to the robustness of Transformer models towards quantization and because \ours performs compression post-adaptation rather than simultaneously with it. 

In general, \ours can maintain the original accuracy of the model most effectively for larger downstream datasets (MNLI, QQP, QNLI, SST2), with an average loss in accuracy of 0.6 points per task. In contrast, the smaller tasks can exhibit much larger accuracy drops, as evidenced by the 3.9 point drop on RTE and 5 point drop on CoLA. In practice, we have further found these two tasks exhibit the greatest amount of hyperparameter sensitivity.

By relaxing our compression constraint to 87.5\%, which is equivalent to a dense, uniform 4-bit compression, \ours search finds inference-efficient configurations that increase our average accuracy by a full point. In five tasks, \ours-Search shows no degradation against the dense baseline. When compared to Q-Bert, \ours-Search achieves equivalent compression and greater inference efficiency while matching its accuracy on MNLI. Like \ours Base, \ours Search still exhibits some weaknesses on smaller tasks, with the accuracy gap on RTE remaining 3.2 points. Section ~\ref{subsec:eval_search} for details on the compression configurations.

\begin{table}[t!]
\small
\newcommand{\colspc}{\hspace{1.0em}}
\setlength{\tabcolsep}{3pt}
\centering
    \caption{The dev set results on the GLUE benchmark. Note that size excludes the footprint of the model's embeddings. \ours successfully retains 98.2\% of the dense BERT\textsubscript{base} model. For \ours-Search, the search algorithm guarantees 87.5\% compression of the encoder stack; each task's configuration may overachieve this constraint. *NxMTransformer did not report a QQP result so this was reproduced in \ours. It differs slightly in that \ours performs post-adaptation compression rather than concurrent compression.}
    \label{tab:glue_results}
    \resizebox{\textwidth}{!}{
    \begin{tabular}{@{}lcccccccccccc@{}} \toprule
                                        & Bits  & Sparsity  & MNLI (m/mm)   & SST-2     & QNLI  & QQP   & CoLA & STS-B & MRPC  & RTE   &           \\ 
                                        \cmidrule{2-3}
Samples                                 &               &           & 392k          & 67k       & 108k  & 368K  & 8.5k  & 5.7k  & 3.5k  & 2.5k  &  Size     & Avg.  \\ \midrule
Baseline (BERT\textsubscript{base})     & No            & No        & 84.5/84.8     & 93.0      & 91.3  & 91.3  & 57.5  & 89.0  & 90.8  & 70.0  &  324M     & 83.5  \\ \midrule
NxMTransformer                          & FP16          & Yes       & 82.3/83.4     & 92.3      & 90.4  & 91.3* & 55.3  & 89.3  & 90.8  & 68.6  &  91.1M    &  82.6 \\
Q-Bert                                  & Q4            & No        & 83.9/84.2     & 92.7      & N/A   & N/A   & N/A   & N/A   & N/A   & N/A   &  41M      &  N/A    \\ \midrule
\ours                                   & Q8            & Yes           & 84.0/84.4     & 92.2  & 90.9  & 91.3  & 52.5  & 88.7  & 90.4  & 67.9  & 51.7M   & 82.5      \\
\ours                                   & Q4            & Yes       & 83.5/83.8     & 92.1      & 90.7  & 91.2  & 52.5  & 88.8  & 89.8  & 66.1  &  31.4M    & 82.1  \\ 
\ours -Search                           & Q4/Q8         & Mixed     & 83.9/84.5     & 92.2      & 91.3  & 91.2  & 57.3  & 89.0  & 91.3  & 66.8  &  <41M     & 83.1  \\
    \bottomrule
\end{tabular}}
\end{table}

\subsection{Quantization Alongside Semi-Structured Sparsity}
\label{subsec:eval_quantization}

The quantization mechanism alongside ADMM can have a significant impact on the achieved accuracy when combining the two techniques. In general, the STE-based QAT approach provides the highest accuracy, outperforming the ADMM-Max approach by a full point across the GLUE suite in the aggressive 4-bit configuration (although their performance is quite similar at 8 bits). ADMM-Max in turn provides higher accuracy than ADMM-Dist in the most utile configurations (e.g., when a reasonable group size is permitted). 

\textbf{QAT-STE vs. ADMM Approaches}
We hypothesize the orthogonality of the STE-QAT approach with respect to the ADMM-based sparsification process enables its greater performance than the ADMM counterparts. The sparsification constraint is a much stronger constraint than quantization in that it induces much greater changes in magnitude of weights; sparsifying a parameter eliminates 100\% of its value, while quantization will only have that large of effect if the chosen scale is twice that of the parameter. As such, the penalty parameter needs to be tuned to the needs of the stronger constraint sparsification constraint. Since STE-QAT does not interact with the ADMM regularizer in any way, it avoids this contention entirely.

\textbf{Impact of Group Size on ADMM-Based Approaches.}
Both ADMM-based approaches show improvement by decreasing the number of weights assigned to each group. When quantized to 4-bits and using a single scale per weight matrix, ADMM-Dist outperforms ADMM-Max on all tasks excepting CoLA, where neither model converges to a meaningful solution. However, the accuracy loss for either approach is unacceptable. As the granularity is increased to 64 neurons per scale, ADMM-Max dramatically improves in accuracy while ADMM-Dist sees more mixed results. In a task like CoLA, ADMM-Dist is able to converge fully; in QNLI, accuracy actually degrades slightly. This phenomenon emerges because ADMM-Max is more sensitive to large outlier weights. As granularity increases, the impact of large outlier weights diminishes. In ADMM-Dist the impact of any single weight is mitigated since it can only contribute so much to the Euclidean distance. 

\begin{table}[t]
\small
\centering
    \caption{Comparison of the different quantization approaches at different quantization levels and grouping granularity. ADMM-Dist minimizes the distance between the quantized and un-quantized representations, ADMM-Max minimizes the quantization error of the largest value, and STE utilizes quantization in the forward pass and an STE on the backpropagation of gradients. Width is the bit-depth of the integer. Group size is the number of output neurons that will share a single scale, where N/A means that all parameters in a weight matrix are quantized with the same weight. All configurations are sparsified using ADMM.}
    \label{tab:quant_results}
    \resizebox{\textwidth}{!}{
    \begin{tabular}{@{}lllccccccccc@{}}
    \toprule
    Method      & Width & Group Size    & MNLI (m/mm)   & SST-2 & QNLI  & QQP   & CoLA  & STS-B & MRPC  & RTE  & Average    \\
    \midrule
    ADMM-Dist   & 8     & N/A           & 83.9/84.4     & 91.7  & 90.6  & 91.2  & 54.4  & 88.8  & 90.0  & 64.3  & 82.2      \\
    ADMM-Max    & 8     & N/A           & 84.1/84.6     & 92.5  & 90.8  & 91.3  & 53.2  & 88.9  & 90.2  & 67.9  & 82.6      \\
    STE-QAT     & 8     & N/A           & 84.0/84.4     & 92.2  & 90.9  & 91.3  & 52.5  & 88.7  & 90.4  & 67.9  & 82.5      \\
    \midrule
    ADMM-Dist   & 4     & N/A           & 79.4/80.1     & 82.5  & 85.9  & 87.8  & 0     & 75.8  & 0.0   & 47.3  & 59.9  \\
    ADMM-Max    & 4     & N/A           & 72.7/73.4     & 73.6  & 82.9  & 81.9  & 10.6  & 54.6  & 0.0   & 47.3  & 55.2  \\
    \midrule
    ADMM-Dist   & 4     & 64            & 80.8/80.6     & 86.3  & 85.4  & 50.6  & 50.6  & 87.7  & 88.3  & 62.5  & 79.0  \\
    ADMM-Max    & 4     & 64            & 81.5/82.9     & 92.1  & 90.4  & 90.2  & 52.7  & 87.8  & 88.7  & 59.2  & 80.5  \\
    \midrule
    ADMM-Max    & 4     & 32            & 82.7/83.6     & 91.4  & 90.4  & 90.3  & 53.8  & 88.2  & 88.5  & 61.4  & 81.1      \\
    STE-QAT     & 4     & 32            & 83.5/83.8     & 92.1  & 90.7  & 91.2  & 52.5  & 88.8  & 89.8  & 66.4  & 82.1      \\
    \bottomrule
    \end{tabular}}
\end{table}

\subsection{Inference Efficiency}
\label{subsec:inference_efficiency}

Since reducing activation precision below 8-bits results in unacceptable accuracy degradation, even models with weight representations at lower precisions are up-casted to the same datatype as the activations for computation in the ALUs. In practice, this means speedup from reducing the data width on its own diminishes. In contrast, N:M semi-structured sparsity relieves ALU pressure by only computing half the operations compared to dense computation. These two approaches to improving performance are complimentary and build on each other.

Using NVIDIA's TensorRT runtime, the combination of both N:M sparsity and integer quantization achieve up to 2.5x end-to-end inference acceleration on BERT\textsubscript{Large} (See Table ~\ref{tab:performance} for full results). In this configuration, quantization alone is able to increase performance by 1.75x, while the semi-structured sparsity improves the quantized performance by 1.42x, demonstrating the importance of considering both techniques for acceleration. In general, more demanding inference configurations (i.e. longer sequence lengths and larger batch sizes) increase the performance benefit of \ours. This is an expected result since less demanding serving configurations are less likely to fully saturate a large GPU like A100. BERT\textsubscript{Large} was chosen as our evaluation model since it represents a compromise between smaller, early Transformers such as BERT\textsubscript{Base} and larger, recent models such as GPT-3. We evaluate at 8-bits only since TensorRT does not yet support 4-bit weights alongside 8-bit activations; the reduced memory bandwidth requirements provide further opportunity for improved performance.

\begin{table}[t]
\small
\centering
 \caption{Measured performance in TensorRT for BERT\textsubscript{Large} across difference sequence lengths and batch sizes on A100 for SQuAD 1.1 ~\cite{performance-results}.}
 ~\label{tab:performance}
 \resizebox{\linewidth}{!}{
    \begin{tabular}{@{}llcccccc@{}} \toprule
    Sequence Length        &   \multicolumn{3}{c}{128} & \multicolumn{3}{c}{384} \\
    \cmidrule{2-4} \cmidrule{5-7}
    Batch Size  & Dense FP16    & Dense INT8 & N:M INT8  & Dense FP16    & Dense INT8 & N:M INT8  \\
    \midrule
    1           & 1.73          & 1.24       & 1.17      & 3.48          & 2.80       & 1.61 \\
    32          & 14.59         & 7.98       & 6.33      & 44.78         & 25.56      & 18.41 \\
    64          & 27.85         & 14.56      & 11.24     &  86.77       & 49.60      & 35.71    \\
    128         & 54.12         & 28.03      & 21.06     & 170.08        & 97.06      & 68.19    \\
    \bottomrule
    \end{tabular}
    }
\end{table}

\subsection{Heterogeneous Compression Configurations}
\label{subsec:eval_search}

\begin{table}[ht]
\small
\centering
    \vspace{-3mm}
    \caption{The chosen compression configuration for each GLUE task, compression ratio, and the reduction in effective FLOPs at equivalent compression as compared to uniform 4-bit integer quantization with a constraint of 87.5\% compression ratio and $K$ equal to ten.}
    \label{tab:compression_configs}
    \resizebox{\textwidth}{!}{
    \begin{tabular}{@{}lcccccccc@{}}
    \toprule
    Task    & Query & Key   & Value & Attention Output  & FFN1      & FFN2  & Compression Ratio & FLOP Reduction \\
    \midrule
    MNLI    & Q4    & Q4        & Q4    & Q4                & Sparse-Q4 & Sparse-Q4 & 89.6\%    & 33.3\%                \\
    SST-2   & Q4    & Q8        & Q4    & Sparse-Q4         & Q4        & Sparse-Q4 & 87.8\%    & 20.1\%                \\
    QNLI    & Q4    & Q8        & Q4    & Sparse-Q4         & Q4        & Sparse-Q4 & 87.8\%    & 20.1\%                \\
    QQP     & Q4    & Q4        & Q4    & Q4                & Sparse-Q4 & Sparse-Q4 & 89.6\%    & 33.3\%                       \\
    CoLA    & Q4    & Q8        & Q4    & Q4                & Q4        & Sparse-Q4 & 87.5\%    & 16.7\&                \\
    STS-B   & Q4    & Sparse-Q4 & Q4    & Q4                & Sparse-Q4 & Sparse-Q4 & 89.8\%    & 37.5\%                \\
    MRPC    & Q4    & Sparse-Q4 & Q4    & Q4                & Sparse-Q4 & Sparse-Q4 & 89.8\%    & 37.5\%                \\
    RTE     & Q4    & Sparse-Q4 & Q4    & Sparse-Q8         & Q4        & Sparse-Q4 & 88.5\%    & 25\%                  \\
    \bottomrule
    \end{tabular}}
\end{table}

\ours-Search is evaluated with a compression constraint of 87.5\% and with $K$ as ten. This enforces the discovered configurations provide at least equal compression as compared with uniform quantization while enabling the flexibility to increase inference efficiency. As seen in Table ~\ref{tab:glue_results}, \ours-Search retains 99.5\% of the baseline's accuracy. The discovered configurations decrease flops by 16.7\% to 37.5\% while reducing the encoder footprint by up to 20\%.

\textbf{Compression Choice Patterns}
While \ours-Search does not choose the same configuration for more than any two tasks, there are some clear patterns in its behavior. For one, the query and value components are never chosen to be sparsified. In fact, across all 31 configurations that were in the top-ten for any given model, just five elected to sparsify the query matrix (one of which up-casted to 8-bit) and 6 for the value matrix (two of which upcasted to 8-bits). In contrast, the second feed-forward layer is sparsified at 4-bits in every retained configuration. This component, which is a third of the footprint of each encoder, appears to extremely amenable as it projects down from the intermediate dimension. The most enigmatic component is the key matrix, which is further sparsified in three configurations and up-casted to 8 bits in three others.

%% file: limitations.tex
\section{Limitations and Potential Negative Social Impacts}
\label{sec:limitations}

\ours is only evaluated on NLU tasks. In addition to other domains of NLP, Transformers have gained increased usage for image tasks as well. Further investigation as to the robustness of this approach to different domains would be valuable. Additionally, Bert\textsubscript{base} is a relatively small model compared to other Transformers and further research into the scalability would also prove beneficial. Finally, since no runtime exists for mixing low-width weights with 8-bit activations, \ours relies on FLOP and compression proxies rather than end-to-end results in all cases.

Transformer models trained on large webtext corpus can embed biases against disadvantaged groups, further discriminating against these groups when deployed in real-world applications. Furthermore, compression of models that does not retain all model accuracy can disproportionately affect discriminated groups. While \ours does not train with a web corpus, its parent models do and \ours will not mitigate those effects.

%% file: conclusion.tex
\section{Conclusion}
\label{sec:conclusion}

In this work, we present \ours, a flexible, compression framework for combining N:M semi-structured sparsity and low-width integer quantization to achieve high levels of inference-friendly compression. We identify QAT-STE as an effective quantization mechanism to work along-side ADMM-based sparsification. Furthermore, we present \ours-Search, an inexpensive, heuristic driven mechanism for heterogeneously compressing Transformers. Evaluated on a wide range of NLU tasks, we find \ours to successfully compress models to 4-bits alongside semi-structured sparsity effectively, while \ours-search can provide inference benefits over homogeneous compression schemes with minimal performance degradation from the parent model.


%% file: appendix.tex
\label{appendix:decomposition}

In this section, we present the full decomposition of the constrained optimization problem via ADMM to Equations~\ref{eq_first-subproblem} and~\ref{eq_second-subproblem}. Using the notation of the paper, the formal constrained optimization problem is given by:

\begin{equation}
    \label{appendix:eq_full_statement}
    \begin{aligned}
    & \underset{\mathbf{W}_i}{minimize} & & f(\{\mathbf{W}_i\}_{i=1}^{L})  \\
    & \text{subject to}  & & \textbf{W}_i \in \textbf{C}
    \end{aligned}
\end{equation}

\noindent where $\textbf{C}$ is some arbitrary compression constraint. To remove the constraint, we introduce an indicator function $g$ (Equation~\ref{appendix:eq_indicator_function}) that returns 0 if the constraint is met and infinity otherwise.

\begin{equation}
    \label{appendix:eq_indicator_function}
      g_i(\mathbf{W}_i) = \begin{cases} 0 & \text{if $\mathbf{W}_i \in \mathbf{C}$} \\ \infty & \text{otherwise} \end{cases} \hspace{10pt} \\
\end{equation}

\noindent The indicator function allows us to reformulate Equation ~\ref{appendix:eq_full_statement} as the following convex optimization problem:

\begin{equation} \label{appendix:eq_reformulation}
\begin{aligned}
& \underset{\mathbf{W}_i}{minimize} & & f(\{\mathbf{W}_i\}_{i=1}^{L}) + \sum_{i=1}^{K} g_i(\mathbf{Z}_i) \\
& \text{subject to}  & & \mathbf{W}_i = \mathbf{Z}_i, i=1,2,...,L
\end{aligned}
\end{equation}

\noindent Equation~\ref{appendix:eq_reformulation} can be decomposed by the augmented Lagrangian, giving Equations~\ref{eq_first-subproblem} and~\ref{eq_second-subproblem}.